\documentclass[conference,a4paper,10pt]{IEEEtran}
\IEEEoverridecommandlockouts
\usepackage[utf8]{inputenc}
\usepackage{times}
\usepackage{graphicx}
\usepackage{amsmath}
\usepackage{amsfonts}
\usepackage{amsthm}
\usepackage{amssymb}
\usepackage{nicefrac}
\usepackage{psfrag}
\usepackage{multirow}
\usepackage{wasysym}
\usepackage{enumerate}
\usepackage[dvipsnames,usenames]{xcolor}
\usepackage{algorithm, algorithmicx, algpseudocode}
\usepackage{url}
\usepackage{bm}
\usepackage{siunitx}
\usepackage{booktabs}

\usepackage{xspace}
\usepackage{booktabs}
\usepackage{tabularx}
\usepackage{fontawesome}
\usepackage{comment}

\usepackage{makecell}
\usepackage{collcell}
\usepackage{etoolbox}
\usepackage{tikz}
\usepackage{adjustbox}
\usepackage{multirow}
\usepackage{pgffor}
\usepackage{pgfmath}
\usepackage{pgf}
\usepackage{numprint} 
\usepackage{pgfplotstable}
\usepackage{colortbl}
\usepackage{color}
\usepackage{pgfkeys}
\usepackage{placeins}
\usepackage{caption}
\usepackage{svg}
\usepackage{xspace}

\usepackage{nccmath}

\pgfplotsset{compat=1.18}
\usepackage[table]{xcolor} 

\ifCLASSOPTIONcompsoc
\usepackage[caption=false, font=normalsize, labelfont=sf, textfont=sf]{subfig}
\else
\usepackage[caption=false, font=footnotesize]{subfig}
\fi

\title{
SVD-Based Typicality Maps for Out-of-Distribution Detection in Vision Transformers}
 \author{
 	\IEEEauthorblockN{
            Aldo Sean Sartor\IEEEauthorrefmark{1},
            Leandro de Souza Rosa\IEEEauthorrefmark{1},
     Andriy Enttsel\IEEEauthorrefmark{1}\thanks{The work of Andriy Enttsel was done before joining Mitsubishi Electric R\&D Centre Europe.}, 
   		Mauro Mangia\IEEEauthorrefmark{1}\IEEEauthorrefmark{2},
        Riccardo Rovatti\IEEEauthorrefmark{1}\IEEEauthorrefmark{2}
        }
 		
 		\IEEEauthorblockA{\IEEEauthorrefmark{1}DEI, \IEEEauthorrefmark{2}ARCES, University of Bologna, Italy; contact author: aldosean.sartor@unibo.it} 
}
\begin{document}

\maketitle

\begin{abstract}
We present a method for analyzing the internal representations of
Vision Transformers (ViTs) exploiting the geometry of their learned parameters.
Each affine layer's weight matrix is factored via Singular Value Decomposition (SVD), and activations are projected onto the leading right singular vectors to obtain compact, layer-intrinsic representations.
A class-conditional density model is then fitted at each layer,
producing per-class \emph{typicality scores} that are stacked across depth into \emph{typicality maps}: two-dimensional summaries of how class-specific evidence evolves through the network.
From these maps, we derive two post-hoc scores for Out-Of-Distribution (OOD) detection: a \emph{Prototype Alignment Score} (PAS), measuring agreement with class reference prototype patterns, and a \emph{Multi-Layer Soft Voting} (MLSV) score, capturing cross-layer consensus without stored prototypes.
On ViT-B/16 fine-tuned on CIFAR-100, the proposed scores achieve
competitive detection performance without retraining or OOD exposure.

\end{abstract}
\begin{IEEEkeywords}
Singular Value Decomposition, Vision Transformer, Out-of-Distribution Detection, Feature Geometry.
\end{IEEEkeywords}




\section{Introduction}

Deep neural networks construct high-dimensional internal
representations through cascades of learned linear transformations and
nonlinearities.
While these representations are central to a model's performance,
their geometry and evolution across depth remain difficult to
characterize, especially for modern deep architectures.
In this work, we focus on Vision Transformer (ViT) \cite{vit} classifiers, where information is progressively transformed by alternating self-attention and multi-layer perceptron (MLP) blocks acting on token embeddings.
A principled, layer-wise view of how class-relevant structure emerges is important both for interpretability and for reliability tasks such as detecting inputs that fall outside the training distribution \cite{yang2024generalized}.

An increasing number of \emph{post-hoc} methods address these questions by analyzing intermediate activations of a frozen, pretrained network, rather than relying solely on the output-layer's
scores \cite{hendrycks2017baseline,liu2020energy}.
Methods based on distances in activation space, such as Deep
Mahalanobis Distance (DMD) \cite{lee2018mahalanobis} and Deep
$k$-Nearest Neighbors ($k$-NN) \cite{sun2022out}, capture distributional information beyond the logits. However, they typically operate on a single layer or require
task-specific supervision, such as Out-Of-Distribution (OOD) validation data for the logistic-regression step in DMD.

More recently, a multi-layer confidence framework called MACS (Multi-layer Analysis for Confidence Scoring)~\cite{capelli2025mlcs} has shown that
aggregating layer-wise evidence into structured \emph{maps} can unify confidence estimation, OOD detection, and adversarial-attack detection.
Concretely, MACS probes the internal activations of a trained network via SVD-based projections \cite{Bishop_2006PRML} and aggregates layer-wise information into structured maps, which are then compared against class-specific prototypes. While MACS demonstrates the effectiveness of this representation for confidence and reliability assessment, it relies on unsupervised clustering in the projected activation space, followed by an empirical feature-label association step to obtain layer-wise class estimates.

\begin{figure}[t]
  \centering
    \includegraphics[width=0.5\textwidth]{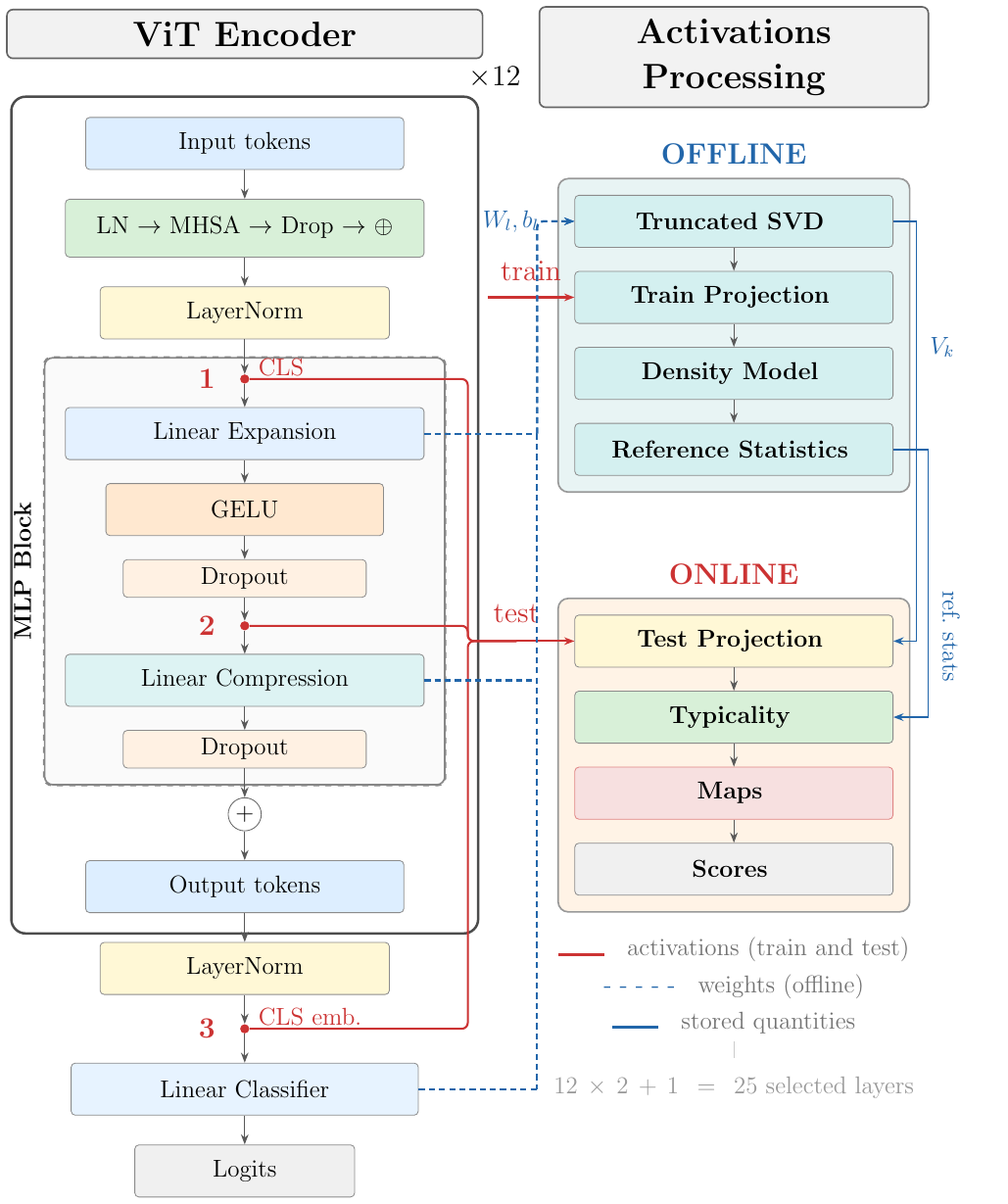}
  \caption{ 
    Overview of the proposed pipeline.
    \textit{Left:} schematic of a ViT encoder block (repeated $\times 12$)
    with the classification head.
    Red dots mark the three hook types used to intercept CLS-token activations.
    \textit{Right:} the activations processing pipeline, divided into
    an \textsc{offline} stage (teal, performed once per model) and an
    \textsc{online} stage (orange, applied to each test sample).
  }
  \label{fig:pipeline}
\end{figure}

In this work, we propose a simpler and probabilistic alternative. We preserve the same geometric backbone: we interpret each affine layer as a linear operator, factorize its augmented weight matrix via Singular Value Decomposition (SVD) \cite{Golub_2023}, and project activations onto the directions along which the layer maximally amplifies its input, to obtain compact, layer-intrinsic representations. However, instead of relying on clustering and empirical feature-label association, we adopt class-conditional density modeling in the SVD-projected space.
This yields per-layer, per-class typicality scores, which, when stacked across depth, form typicality maps that play the same structural role as MACS classification maps while reducing the number of processing stages and hyperparameters and providing scores with a direct probabilistic interpretation. These maps reveal how class-specific evidence emerges through the network.
From these maps, we then derive two scalar scores: the \emph{Prototype Alignment Score (PAS)}, measuring agreement with class-specific reference maps analogous to MACS's prototype-matching; 
and the \emph{Multi-Layer Soft Voting (MLSV)} score, which captures cross-layer consensus without requiring any stored prototypes. Both scores are fully post-hoc and do not require access to OOD data.

 We validate the framework on ViT-B/16 fine-tuned on CIFAR-100.
PAS and MLSV achieve competitive OOD detection against established
baselines while providing a depth-resolved
representation of the network's internal dynamics.

Our contributions can be summarized as follows:
\begin{itemize}
    \item \textbf{Probabilistic reformulation}: We replace MACS's clustering and feature-label association with class-conditional density modeling in the SVD-projected space, yielding per-layer, per-class typicality scores with a direct probabilistic interpretation.
    
    \item \textbf{Simpler maps}: We introduce typicality maps, which play the same role as MACS classification maps but with a simpler pipeline and fewer hyperparameters.
    
    \item \textbf{Novel scores}: We derive two post-hoc scores: PAS, which simplifies MACS's prototype-matching score, and MLSV, a new prototype-free score capturing cross-layer consensus.
\end{itemize}

The remainder of this paper is organized as follows. Section~\ref{sec:models} presents the mathematical models. Section~\ref{sec:setup} defines the experimental setup and describes the datasets. Section~\ref{sec:Results} reports and discusses the empirical findings. Section~\ref{sec:conclusions} concludes the paper.

\section{Mathematical Model}
\label{sec:models}

The proposed method consists of five stages: \textit{i}) SVD-based projection of activations, \textit{ii}) class-conditional density modeling in the projected space, \textit{iii}) computation of per-layer typicality scores, \textit{iv}) aggregation of these scores into depth-resolved maps, and \textit{v}) scalar score computation.

\subsection{SVD-Based Projection of Internal Activations}

Consider a neural network, out of which $L$ affine candidate layers are selected. Focusing on the transformation performed by a specific layer $\ell \in \{1, \dots {L} \}$, given the input activation $x_\ell \in \mathbb{R}^{d_\ell}$, the output is
\begin{equation}
y_\ell = W_\ell x_\ell + b_\ell, \qquad W_\ell \in \mathbb{R}^{m_\ell \times d_\ell}, \; b_\ell \in \mathbb{R}^{m_\ell}.
\end{equation}
As in MACS~\cite{capelli2025mlcs}, we rewrite this affine transformation using an augmented input:
\begin{equation}
\tilde{x}_\ell = 
\begin{bmatrix}
x_\ell \\ 1
\end{bmatrix}
\in \mathbb{R}^{d_\ell+1}, 
\qquad
A_\ell = [\, W_\ell \;\; b_\ell \,] \in \mathbb{R}^{m_\ell \times (d_\ell+1)},
\end{equation}
so that $y_\ell = A_\ell \tilde{x}_\ell$.

We then compute the SVD \cite[Ch.~2]{Golub_2023} of $A_\ell$:
\begin{equation}
A_\ell = U_\ell \Sigma_\ell V_\ell^\top,
\end{equation}
where $U_\ell \in \mathbb{R}^{m_\ell \times m_\ell}$ and $V_\ell \in \mathbb{R}^{(d_\ell+1) \times (d_\ell+1)}$ are orthonormal matrices, and $\Sigma_\ell$ contains the singular values in non-increasing order.
This allows us to define the projected representation at layer $\ell$ as
\begin{equation}
\label{eq: projection}
z_\ell = V_{\ell,k}^\top \tilde{x}_\ell \in \mathbb{R}^{k},
\end{equation}
where $V_{\ell,k} \in \mathbb{R}^{(d_\ell+1) \times k}$ denotes the matrix formed by the first $k$ right singular vectors. 
The projection expresses the activation in a coordinate system induced by the layer’s parameters, aligned with the principal directions of the affine operator $A_\ell$. The dimensionality $k$ is chosen such that $k \ll d_\ell+1$, yielding a compact, geometry-aware representation.

\subsection{Class-Conditional Density Modeling}

Let $\mathcal{D}_{\text{train}} = \{(x^{(j)}, y^{(j)})\}_{j=1}^N$ be the training set, where $y^{(j)} \in \{1, \dots, {C} \}$ denotes the class label. For each selected layer $\ell$, we compute the projected representations $z_\ell^{(j)}$ using \eqref{eq: projection}.
Then for each layer $\ell$ and class $c$, we fit a class-conditional probability density $p_{\ell,c}(z)$ to the set $\{ z_\ell^{(j)} \mid y^{(j)} = c \}$. Each density is modeled using a Gaussian Mixture Model (GMM)~\cite[Ch.~9]{Bishop_2006PRML} with $M$ components:
\begin{equation}
p_{\ell,c}(z) = \sum_{m=1}^{M} \pi_{\ell,c,m} \, \mathcal{N}(z; \mu_{\ell,c,m}, \Sigma_{\ell,c,m}),
\end{equation}
where $\pi_{\ell,c,m}$ are the mixture weights and $(\mu_{\ell,c,m}, \Sigma_{\ell,c,m})$ are the mean and the covariance of each component.

This replaces MACS’s clustering and association pipeline with a single density modeling step, yielding quantities that allow a direct likelihood-based interpretation.

\subsection{Typicality Scores}

Given a test sample $x$, we obtain its projected representation $z_\ell(x)$ at each layer $\ell$. For a class $c$, we first evaluate the negative log-likelihood:
\begin{equation}
s_{\ell,c}(x) = - \log p_{\ell,c}( z_\ell(x) ).
\end{equation}
Then, to obtain a scale-independent and comparable measure across layers and classes, we normalize these scores using the Empirical Cumulative Distribution Function (ECDF) estimated on the training set for each $(\ell, c)$ pair: 
\begin{equation}
\widehat{F}_{\ell,c}(t) = \frac{1}{N_{c}} \sum_{j: y^{(j)}=c} \mathbf{1}\big[ s_{\ell,c}(x^{(j)}) \le t \big],
\end{equation}
where $N_{c}$ is the number of training samples of class $c$. 

With this, we define the typicality score as:
\begin{equation}
\tau_{\ell,c}(x) = 1 - \widehat{F}_{\ell,c}\big( s_{\ell,c}(x) \big) \in [0,1].
\end{equation}
High values of $\tau_{\ell,c}(x)$ indicate that the representation of $x$ at layer $\ell$ is typical for class $c$, while low values indicate atypical or unlikely behavior. These scores play the same conceptual role as the layer-wise class estimates in MACS, but are obtained through probabilistic modeling and likelihood evaluation rather than clustering and association.

\subsection{Typicality Maps and Class Prototypes}

For a given input $x$, we collect all typicality scores into a matrix, called the typicality map:
\begin{equation}
T(x) \in [0,1]^{C \times L}, 
\qquad
T(x)_{c,\ell} = \tau_{\ell,c}(x).
\end{equation}

Each row of $T(x)$ quantifies how typical the sample is with respect to a fixed class across depth, while each column describes the class-wise typicality profile at a given layer.
Structurally, $T(x)$ is analogous to the classification-map in MACS: both encode a depth-resolved, class-wise summary of the network’s internal decision process.

Similarly to MACS proto-maps, we define a  \emph{prototype map}  for each class $c$ by averaging the typicality maps of correctly classified training samples:

\begin{equation}
\overline{T}_c = \frac{1}{|\mathcal{P}_c|} \sum_{x \in \mathcal{P}_c} T(x),
\end{equation}
where
\begin{equation}
\mathcal{P}_c = \{ x \in \mathcal{D}_{\text{train}} \mid y(x) = \hat{y}(x) = c \}.
\end{equation}
These prototype maps represent the expected depth-wise typicality pattern for each class.

\subsection{Scalar Scores}
\label{Scalar_scores}

The map structure allows us to define two compact scalar scores:
\paragraph{Prototype Alignment Score (PAS)}

Given a test sample $x$ with predicted class $\hat{y}(x)$, we measure the alignment between its map and the corresponding class prototype using the Frobenius scalar product:
\begin{equation}
S_{\text{PAS}}(x) = 
\frac{ \langle T(x), \overline{T}_{\hat{y}(x)} \rangle_F }
{ \| T(x) \|_F \, \| \overline{T}_{\hat{y}(x)} \|_F } \in [0,1],
\end{equation}
where $\langle A, B \rangle_F = \sum_{i,j} A_{i,j} B_{i,j}$ and $\| \cdot \|_F$ denotes the Frobenius norm. High values indicate strong agreement with the typical class-specific pattern. This score is a probabilistic reformulation of the MACS scoring function, which compares a classification-map to a class proto-map using the cosine similarity.

\paragraph{Multi-Layer Soft Voting (MLSV)}
First, we normalize each column of $T(x)$ with a softmax across classes:
\begin{equation}
\tilde{\tau}_{\ell,c}(x) = 
\frac{ \exp( \tau_{\ell,c}(x) ) }
{ \sum_{c'=1}^{C} \exp( \tau_{\ell,c'}(x) ) }.
\end{equation}
We then aggregate these normalized scores across layers and apply a second softmax:
\begin{equation}
s_c(x) = 
\frac{ \exp\!\left( \sum_{\ell=1}^{L} \tilde{\tau}_{\ell,c}(x) \right) }
{ \sum_{c'=1}^{C} \exp\!\left( \sum_{\ell=1}^{L} \tilde{\tau}_{\ell,c'}(x) \right) }.
\end{equation}
The MLSV score is defined as
\begin{equation}
S_{\text{MLSV}}(x) = \max_{c} s_c(x) \in \left[ \tfrac{1}{C}, 1 \right].
\end{equation}
High values indicate strong cross-layer consensus toward a single class, while low values indicate disagreement across layers.

The overall five-stage pipeline is summarized in (Fig.~\ref{fig:pipeline}, right), where we also distinguish between the offline and online phases.

\section{Experimental Setup}
\label{sec:setup}

\subsection{Dataset and Model}
We fine-tune a ViT-B/16 pretrained on
ImageNet-1k~\cite{deng2009imagenet} on the CIFAR-100
training set~\cite{cifar}.
The resulting model achieves $86.6\%$ top-1 test accuracy.

We apply the proposed framework to the MLP sub-blocks of each ViT
encoder layer.
Specifically, we intercept CLS-token activations at two points within
each MLP: before the linear expansion and before the linear compression
(Fig.~\ref{fig:pipeline}, left), yielding $2 \times 12 = 24$
intermediate hook points.
An additional hook is placed at the input of the classification head,
for a total of $L = 25$ monitored layers.

For each hooked layer, activations are projected onto the top-$k$
right singular vectors of the corresponding weight matrix.
The projection dimension is fixed to $k = 200$ for intermediate
layers and $k = 100$ for the classification head, matching the number
of classes.

Class-conditional densities are modeled by GMMs
with $M = 4$ diagonal-covariance components per class and covariance
regularization $\epsilon = 10^{-4}$.
Unlike MACS, which models the joint representation
space, our approach operates in a class-conditional setting, allowing a significantly lower number of mixture components.
All hyperparameters \{$k$, $M$, $\epsilon$\} were selected empirically
in preliminary experiments, the same configuration is used throughout
and no task-specific tuning was performed.

\subsection{Out-of-Distribution Datasets}
Following~\cite{fort2021exploring} and~\cite{near_far_ood}, we evaluate
on three datasets that are semantically disjoint from CIFAR-100 and
cover different visual domains:
\begin{itemize}
\item SVHN~\cite{svhn}: house-number photographs from
      Google Street View.
\item Places365~\cite{places365}: scene photographs
      depicting indoor and outdoor environments.
\item DTD~\cite{cimpoi14describing}: the Describable Textures Dataset, with
      images of natural textures organized into perceptual attribute
      categories.
\end{itemize}
\section{Results}
\label{sec:Results}
\subsection{Visualizing Maps}

\begin{figure}[!t]
\centering
\includegraphics[width=\columnwidth]{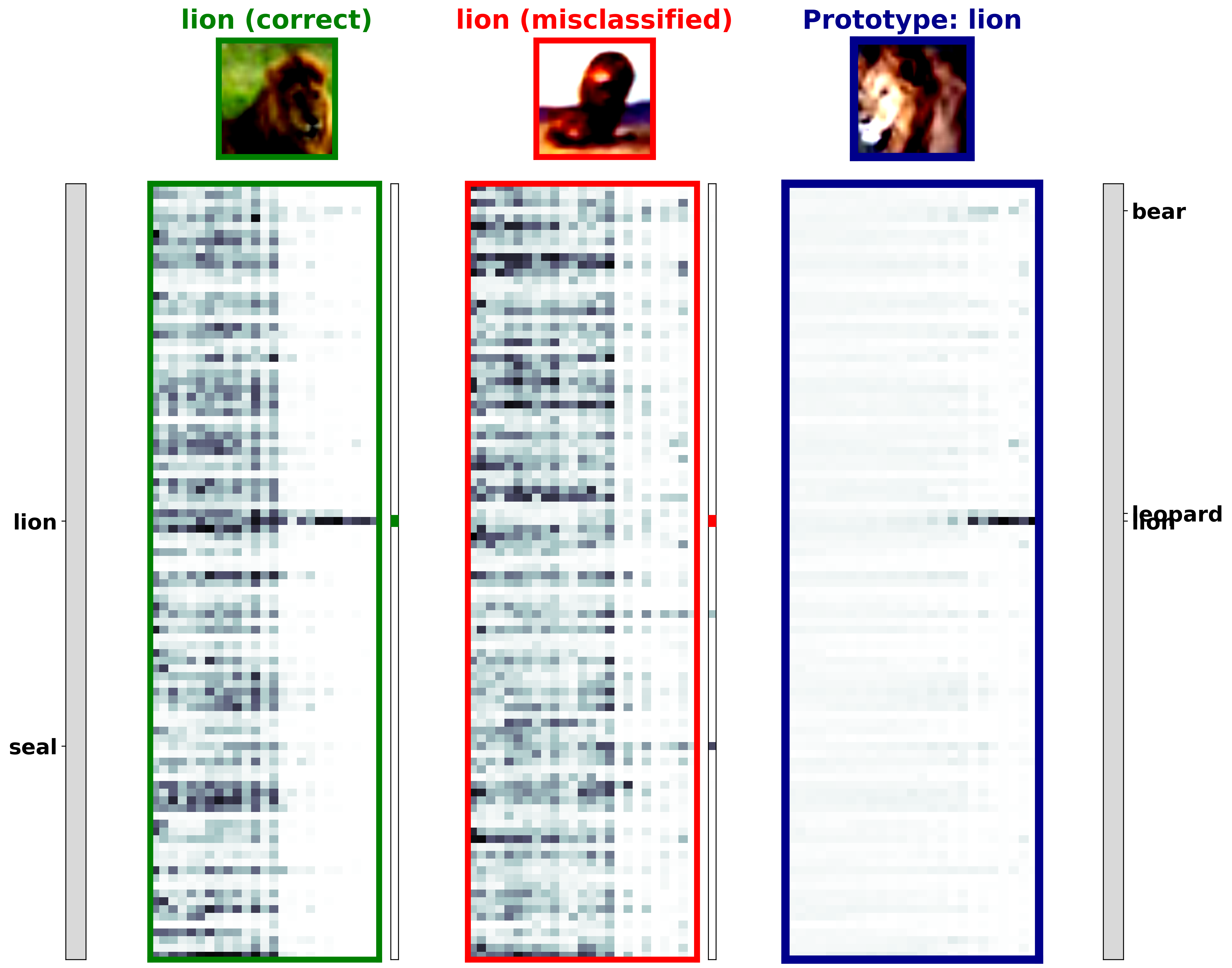}
\caption{Typicality maps for the first correctly classified lion (green border), the first misclassified lion (red border) and the class prototype.
  Left sidebar indicates ground-truth and predicted labels, right sidebar lists the top-3 classes according
  to the prototype.}
\label{fig:maps_protoclass}
\end{figure}

\label{subsec:representation_analysis}
We illustrate the structure of the proposed maps $T(x)$ on a few representative samples in Fig.~\ref{fig:maps_protoclass}. For correctly classified inputs, typicality tends to concentrate on the ground-truth class as depth increases, while in early layers typicality is spread across multiple classes
(Fig.~\ref{fig:maps_protoclass}, green).
By contrast, misclassified samples do not exhibit the same
pattern: typicality remains spread across competing classes or
fluctuates across layers, even in the last layers 
(Fig.~\ref{fig:maps_protoclass}, red).
Averaging maps over correctly classified training samples yields class
prototypes that summarize a typical depth-wise profile; semantically
related classes retain moderate typicality, reflecting inter-class
similarity in the network's representations (Fig.~\ref{fig:maps_protoclass}, blue).

\begin{figure}[!t]
\centering
\includegraphics[width=\columnwidth]{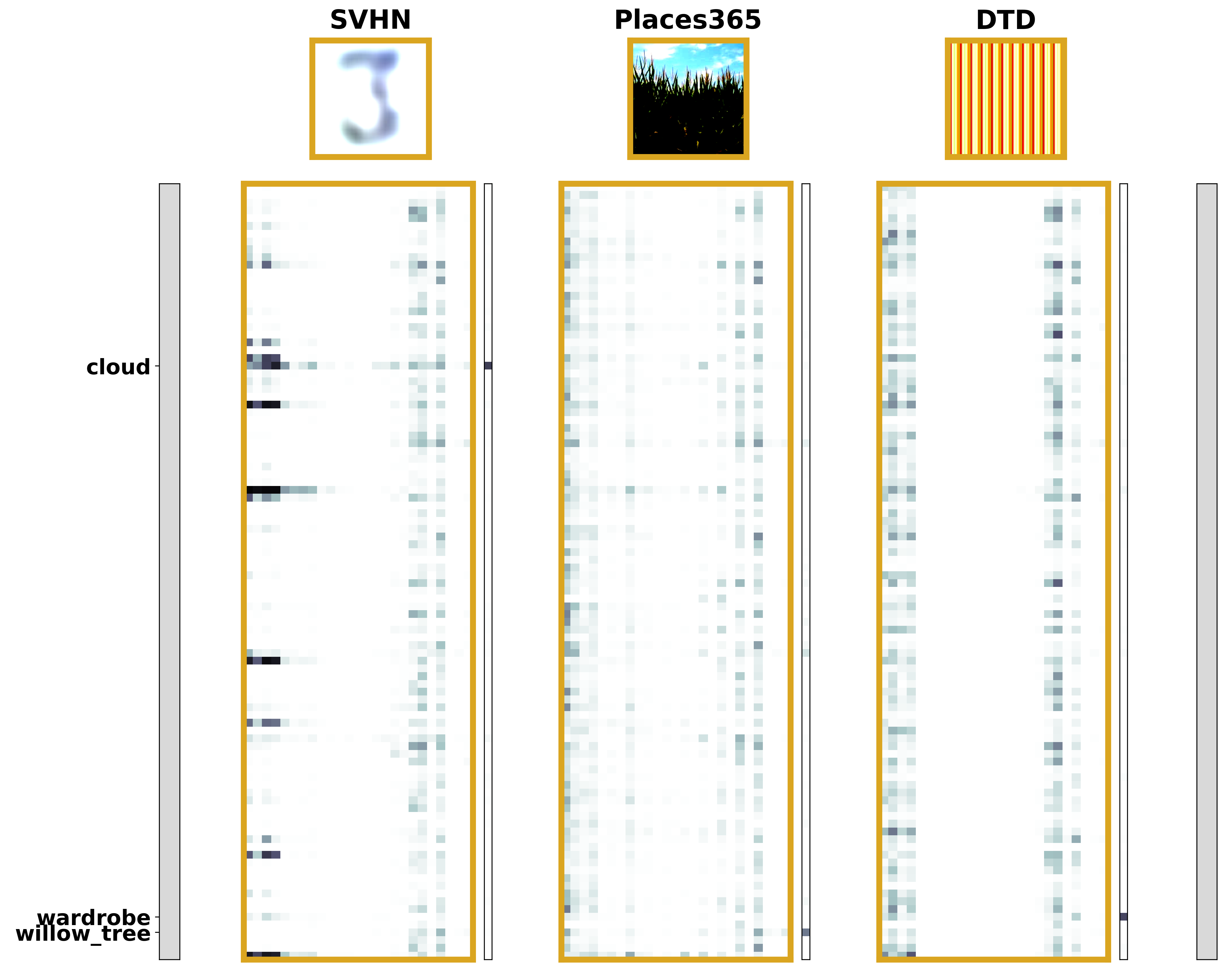}
\caption{Typicality maps for the first samples of each OOD dataset (SVHN, Places365, DTD).
Left sidebar indicates the predicted label.}
\label{fig:maps_ood}
\end{figure}

OOD inputs instead display diffuse, unstructured patterns across all
classes and layers (Fig.~\ref{fig:maps_ood}), consistent with their
representations falling outside the high-density regions of the
class-conditional models.

\subsection{Out-of-Distribution Detection}

We evaluate PAS and MLSV on OOD detection, where
test samples from unseen distributions are expected to produce
less structured maps (cf.\ Fig.~\ref{fig:maps_ood}).

We compare against the following post-hoc detectors:
\emph{MSP}~\cite{hendrycks2017baseline}, the maximum softmax probability; \emph{MLS}~\cite{pmlr-v162-hendrycks22a}, the maximum logit;
\emph{Energy}~\cite{liu2020energy}, the log-sum-exp of logits;
\emph{ReAct}~\cite{sun2021react}, which clips penultimate-layer
activations at the 90th training percentile before computing Energy;
\emph{DMD-B}~\cite{lee2018mahalanobis}, single-layer Mahalanobis
distance computed on the penultimate-layer features, i.e., the input to the classification head; \emph{$k$-NN}~\cite{sun2022out}, the negative distance to the $k$-th nearest
training feature ($k{=}50$) in the $\ell_2$-normalised embedding space;
and \emph{DMD-A}~\cite{lee2018mahalanobis} as an oracle
upper bound, which aggregates per-layer Mahalanobis scores via a logistic regression (LR) fit on OOD validation data.

\begin{table}[!t]
\centering
\caption{OOD detection performance reported as AUROC$\uparrow$ (\%) / FPR@95$\downarrow$ (\%). ID = CIFAR-100, backbone ViT-B/16.}
\label{tab:ood_full}
\resizebox{\columnwidth}{!}{%
\begin{tabular}{lcccc}
\toprule
\textbf{Score}
& \textbf{SVHN}
& \textbf{Places365}
& \textbf{DTD}
& \textbf{Mean} \\
\midrule
MSP        & 89.8 / 44.1          & 83.9 / 59.6          & 92.2 / 33.1          & 88.6 / 45.6 \\
MLS        & 94.9 / 24.2          & 91.9 / 39.9          & 97.2 / 14.2          & 94.6 / 26.1 \\
Energy     & 95.3 / 21.7          & 92.7 / 36.4          & 97.6 / 11.5          & 95.2 / 23.2 \\
ReAct      & 95.1 / 22.4          & 93.3 / 33.3          & 97.6 / 11.2          & 95.3 / 22.3 \\
 $k$-NN       & \textbf{95.9} / 22.4 & 89.6 / 49.7          & 96.7 / 16.6          & 94.1 / 29.6 \\

DMD-B      & 90.5 / 67.6          & 96.9 / 14.6          & 98.6 / \phantom{0}6.6 & 95.3 / 29.6 \\
\rowcolor{black!6}
DMD-A$^\dagger$
  & 99.0 / \phantom{0}3.9 & 99.9 / \phantom{0}0.1 & 99.9 / \phantom{0}0.1 & 99.6 / \phantom{0}1.4 \\

MACS   & 90.3 / 51.4          & 87.6 / 60.4          & 94.1 / 36.1          & 90.6 / 49.3 \\

\midrule
PAS  (ours)  & 94.9 / 31.3          & 90.7 / 44.0          & 96.5 / 19.6          & 94.0 / 31.6 \\
MLSV (ours)  & 92.3 / 31.4          & \textbf{99.4} / \textbf{\phantom{0}1.8} & \textbf{99.4} / \textbf{\phantom{0}2.4} & \textbf{97.0} / \textbf{11.9} \\

\bottomrule
\multicolumn{5}{l}{\footnotesize $^\dagger$\,DMD-A uses OOD
  validation data for LR;
  other methods are agnostic.}
\end{tabular}}
\end{table}

Table~\ref{tab:ood_full} reports AUROC and FPR@95 for all methods.
Among unsupervised detectors, MLSV achieves the highest mean AUROC
$97.0\%$ and the lowest mean FPR@95 $11.9\%$, outperforming the strongest baseline by $1.7$ percentage points (pp) in AUROC
and $10.4$ pp in FPR@95. This advantage is most pronounced on
Places365 and DTD, where MLSV reaches $99.4\%$ AUROC with FPR@95
below $2.5\%$, closely approaching the supervised oracle DMD-A$^\dagger$.

PAS attains a mean AUROC of $94.0\%$, comparable to MLS $94.6\%$
and $k$-NN $94.1\%$, while outperforming MACS ($90.6\%$) by $3.4$ pp in
mean AUROC and $17.7$ pp in mean FPR@95, confirming that the simpler
PAS formulation is a more effective aggregation of the multi-layer
information.

The one exception is SVHN, where logit-based scores (Energy $95.3\%$,
ReAct $95.3\%$, $k$-NN $95.9\%$) outperform both MLSV $92.3\%$ and PAS $94.9\%$. On this dataset the large distributional gap between digit
images and CIFAR-100 categories is captured by simple output
statistics, leaving limited room for map-based methods. Across the
remaining two shifts, however, the ranking reverses: MLSV improves
over Energy by $6.7$ pp on Places365 and $1.8$ pp on DTD in AUROC,
with FPR@95 reductions of $34.6$ pp and $9.1$ pp, respectively.

\begin{figure}[!t]
\centering
\includegraphics[width=\columnwidth]{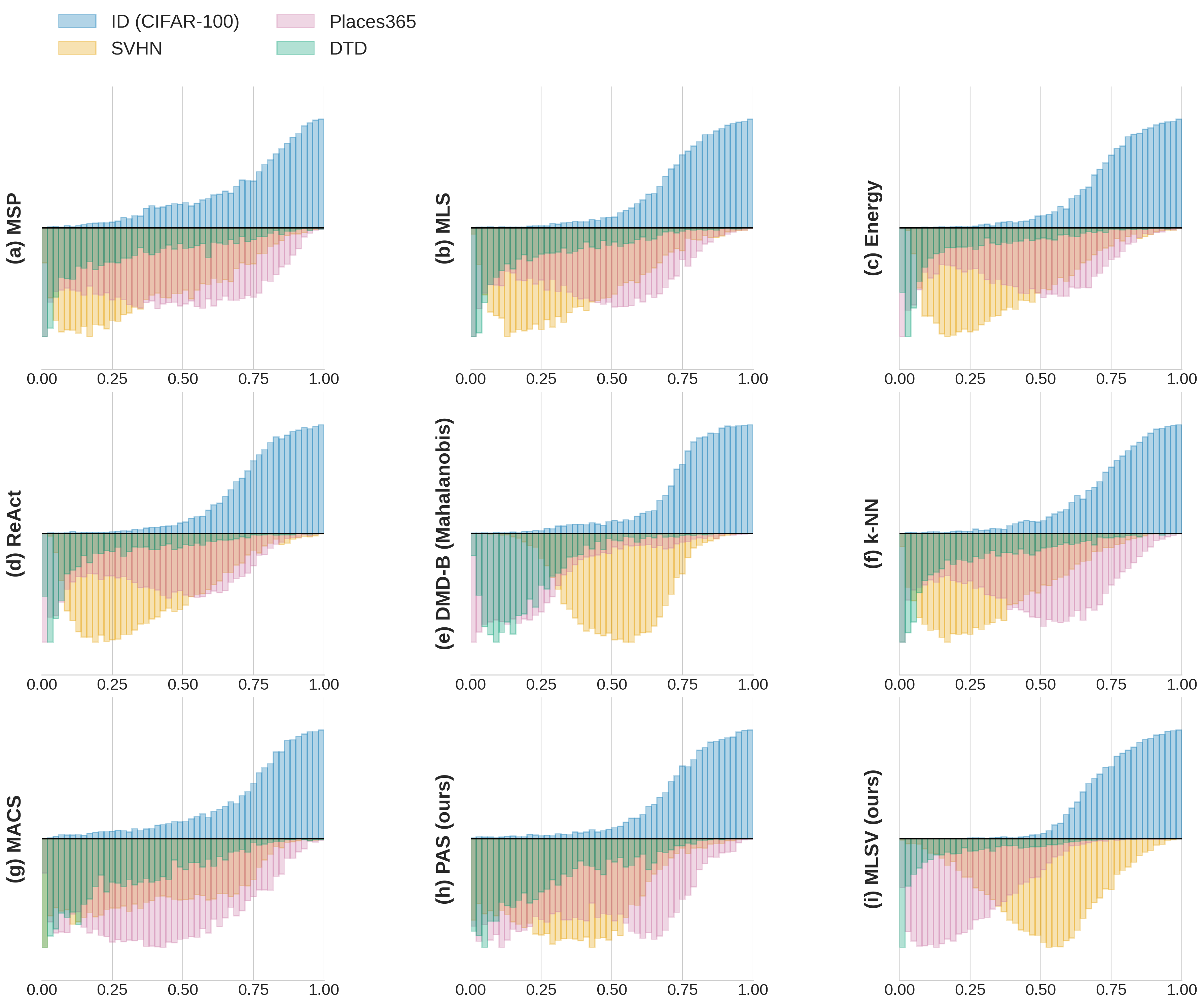}
\caption{Normalized score distributions: in-distribution (CIFAR-100 test, above axis)
  vs.\ OOD (SVHN, Places365, DTD, mirrored below) for all evaluated methods.}
\label{fig:violin}
\end{figure}

Fig.~\ref{fig:violin} provides a distributional view of all
evaluated methods after ECDF normalization and per-histogram
rescaling to unit peak height.
Logit-based scores (MSP, MLS, Energy, ReAct) show moderate
overlap between the ID and OOD distributions, particularly on
Places365 and DTD where the OOD mass shifts only partially
away from the ID peak.
DMD-B and $k$-NN improve separation on specific shifts but remain
inconsistent across datasets.
MACS, PAS, and MLSV progressively reduce this overlap:
MLSV in particular pushes the OOD distributions towards the
low-score tail on all three benchmarks, with the ID distribution
concentrated at high scores, consistent with the AUROC and
FPR@95 results of Table~\ref{tab:ood_full}.

\section{Conclusions}
\label{sec:conclusions}

Building on \cite{capelli2025mlcs}, we presented a framework for
analysing the internal representations of Vision Transformers through
SVD-based projections and class-conditional density modeling.
The resulting typicality maps provide a depth-resolved view of how
class-specific evidence consolidates across layers, and serve as the
basis for two post-hoc OOD detection scores.

PAS measures alignment with stored class prototype maps and
outperforms the closely related MACS score while requiring
significantly fewer parameters.
MLSV discards stored prototypes entirely and instead leverages
cross-layer consensus: by aggregating lightweight per-layer
typicality votes without any stored reference, it achieves the
best results among all unsupervised detectors considered across
all three OOD benchmarks, and approaches the performance of the
supervised oracle on two of them.

The reliance on independent per-layer modelling suggests that
representational consistency across depth is a strong reliability
signal that existing single-layer or logit-based approaches may
not fully exploit.
\bibliographystyle{IEEEtran}
\bibliography{reference}

@techreport{cifar,
  title        = {Learning Multiple Layers of Features from Tiny Images},
  author       = {Krizhevsky, Alex},
  institution  = {University of Toronto},
  year         = {2009}
}

@inproceedings{svhn,
  title     = {Reading Digits in Natural Images with Unsupervised Feature Learning},
  author    = {Netzer, Yuval and Wang, Tao and Coates, Adam and Bissacco, Alessandro and Wu, Bo and Ng, Andrew Y.},
  booktitle = {NIPS Workshop on Deep Learning and Unsupervised Feature Learning},
  year      = {2011}
}

@article{places365,
  title   = {Places: A 10 Million Image Database for Scene Recognition},
  author  = {Zhou, Bolei and Lapedriza, Agata and Khosla, Aditya and Oliva, Aude and Torralba, Antonio},
  journal = {IEEE Transactions on Pattern Analysis and Machine Intelligence},
  volume  = {40},
  number  = {6},
  pages   = {1452--1464},
  year    = {2018}
}

@InProceedings{cimpoi14describing,
  author    = {M. Cimpoi and S. Maji and I. Kokkinos and S. Mohamed and A. Vedaldi},
  title     = {Describing Textures in the Wild},
  booktitle = {Proceedings of the {IEEE} Conf. on Computer Vision and Pattern Recognition ({CVPR})},
  year      = {2014}
}

@inproceedings{deng2009imagenet,
  title={Imagenet: A large-scale hierarchical image database},
  author={Deng, Jia and Dong, Wei and Socher, Richard and Li, Li-Jia and Li, Kai and Fei-Fei, Li},
  booktitle={2009 IEEE conference on computer vision and pattern recognition},
  pages={248--255},
  year={2009},
  organization={{IEEE}}
}

@inproceedings{vit,
  title     = {An Image is Worth 16x16 Words: Transformers for Image Recognition at Scale},
  author    = {Dosovitskiy, Alexey and Beyer, Lucas and Kolesnikov, Alexander and Weissenborn, Dirk and Zhai, Xiaohua and Unterthiner, Thomas and Dehghani, Mostafa and Minderer, Matthias and Heigold, Georg and Gelly, Sylvain and Uszkoreit, Jakob and Houlsby, Neil},
  booktitle = {International Conference on Learning Representations (ICLR)},
  year      = {2021}
}

@article{near_far_ood,
  title   = {Contrastive Training for Improved Out-of-Distribution Detection},
  author  = {Winkens, Jim and Bunel, Rudy and Roy, Abhijit Guha and Stanforth, Robert and Natarajan, Vivek and Ledsam, Joseph R. and MacWilliams, Patricia and Kohli, Pushmeet and Karthikesalingam, Alan and Kohl, Simon and Cemgil, Taylan and Eslami, S. M. Ali and Ronneberger, Olaf},
  journal = {arXiv preprint arXiv:2007.05566},
  year    = {2020}
}

@inproceedings{hendrycks2017baseline,
  title     = {A Baseline for Detecting Misclassified and Out-of-Distribution Examples in Neural Networks},
  author    = {Hendrycks, Dan and Gimpel, Kevin},
  booktitle = {ICLR},
  year      = {2017}
}

@inproceedings{liu2020energy,
  title     = {Energy-based Out-of-distribution Detection},
  author    = {Liu, Weitang and Wang, Xiaoyun and Owens, John D. and Li, Yixuan},
  booktitle = {NeurIPS},
  year      = {2020}
}

@inproceedings{lee2018mahalanobis,
  title     = {A Simple Unified Framework for Detecting Out-of-Distribution Samples and Adversarial Attacks},
  author    = {Lee, Kimin and Lee, Kibok and Lee, Honglak and Shin, Jinwoo},
  booktitle = {NeurIPS},
  year      = {2018}
}

@article{sun2021react,
  title={React: Out-of-distribution detection with rectified activations},
  author={Sun, Yiyou and Guo, Chuan and Li, Yixuan},
  journal={Advances in Neural Information Processing Systems},
  volume={34},
  pages={144--157},
  year={2021}
}

@article{fort2021exploring,
  title   = {Exploring the Limits of Out-of-Distribution Detection},
  author  = {Fort, Stanislav and Ren, Jie and Lakshminarayanan, Balaji},
  journal = {Advances in Neural Information Processing Systems},
  volume  = {34},
  pages   = {7068--7081},
  year    = {2021}
}

@article{yang2024generalized,
  title={Generalized out-of-distribution detection: A survey},
  author={Yang, Jingkang and Zhou, Kaiyang and Li, Yixuan and Liu, Ziwei},
  journal={International Journal of Computer Vision},
  volume={132},
  number={12},
  pages={5635--5662},
  year={2024},
  publisher={Springer}
}

@inproceedings{sun2022out,
  title={Out-of-distribution detection with deep nearest neighbors},
  author={Sun, Yiyou and Ming, Yifei and Zhu, Xiaojin and Li, Yixuan},
  booktitle={International conference on machine learning},
  pages={20827--20840},
  year={2022},
  organization={PMLR}
}

@InProceedings{pmlr-v162-hendrycks22a,
  title = 	 {Scaling Out-of-Distribution Detection for Real-World Settings},
  author =       {Hendrycks, Dan and Basart, Steven and Mazeika, Mantas and Zou, Andy and Kwon, Joseph and Mostajabi, Mohammadreza and Steinhardt, Jacob and Song, Dawn},
  booktitle = 	 {Proceedings of the 39th International Conference on Machine Learning},
  pages = 	 {8759--8773},
  year = 	 {2022},

}

@article{capelli2025mlcs,
  title   = {Multi-Layer Confidence Scoring for Detection of Out-of-Distribution Samples, Adversarial Attacks, and In-Distribution Misclassifications},
  author  = {Capelli, Lorenzo and de Souza Rosa, Leandro and Setti, Gianluca and Mangia, Mauro and Rovatti, Riccardo},
  journal = {arXiv preprint arXiv:2512.19472},
  year    = {2025}
}

@book{Bishop_2006PRML,
  author    = {Christopher M. Bishop},
  title     = {Pattern Recognition and Machine Learning},
  publisher = {Springer},
  address   = {New York, NY, USA},
  year      = {2006},
  isbn      = {978-0-387-31073-2}
}

@book{Golub_2023,
  title     = {Matrix Computations},
  author    = {Golub, Gene H. and Van Loan, Charles F.},
  edition   = {4},
  year      = {2013},
  publisher = {Johns Hopkins University Press}
}
\end{document}